\documentclass{article} 
\usepackage{iclr2024_conference,times}

\usepackage[utf8]{inputenc} 
\usepackage[T1]{fontenc}    
\usepackage{hyperref}       
\usepackage{url}            
\usepackage{booktabs}       
\usepackage{amsfonts}       
\usepackage{nicefrac}       
\usepackage{microtype}      
\usepackage{amsmath}
\usepackage{graphicx}
\usepackage{caption, subcaption} 
\usepackage{setspace}
\usepackage{amssymb, bm, amsthm, amsmath}
\usepackage[linesnumbered, ruled, vlined]{algorithm2e}
\usepackage{xcolor}
\usepackage{multirow}
\usepackage{wrapfig}
\newcommand{\modelnameA}{{\sc BSDetector}} 
\input{def.set}
\usepackage{caption} 
\usepackage{bbm}
\usepackage{lipsum}

\title{Quantifying Uncertainty in Answers from any Language Model and Enhancing their Trustworthiness}

\author{Jiuhai Chen \\
Cleanlab, University of Maryland \\
\texttt{jchen169@umd.edu}
\And
Jonas Mueller \\
Cleanlab \\
\texttt{jonas@cleanlab.ai}
}

\iclrfinalcopy 
\begin{document}

\maketitle

\begin{abstract}
We introduce \modelnameA{}, a method for detecting bad and speculative answers from a pretrained Large Language Model by estimating a numeric confidence score for any output it generated. Our uncertainty quantification technique works for any LLM accessible only via a black-box API, whose training data remains unknown.
By expending a bit of extra computation, users of any LLM API can now get the same response as they would ordinarily, as well as a confidence estimate that cautions when not to trust this response. Experiments on both closed and open-form Question-Answer benchmarks reveal that \modelnameA{} more accurately identifies incorrect LLM responses than alternative uncertainty estimation procedures (for both GPT-3 and ChatGPT). 
By sampling multiple responses from the LLM and considering the one with the highest confidence score, we can additionally obtain more accurate responses from the same LLM, without any extra training steps. In applications involving automated evaluation with LLMs, accounting for our confidence scores leads to more reliable evaluation in both human-in-the-loop and fully-automated settings (across both GPT 3.5 and 4). 
\end{abstract}

\begin{figure}[ht]
     \centering
     \begin{subfigure}[b]{1.0\textwidth}
         \centering
         \includegraphics[width=\textwidth]{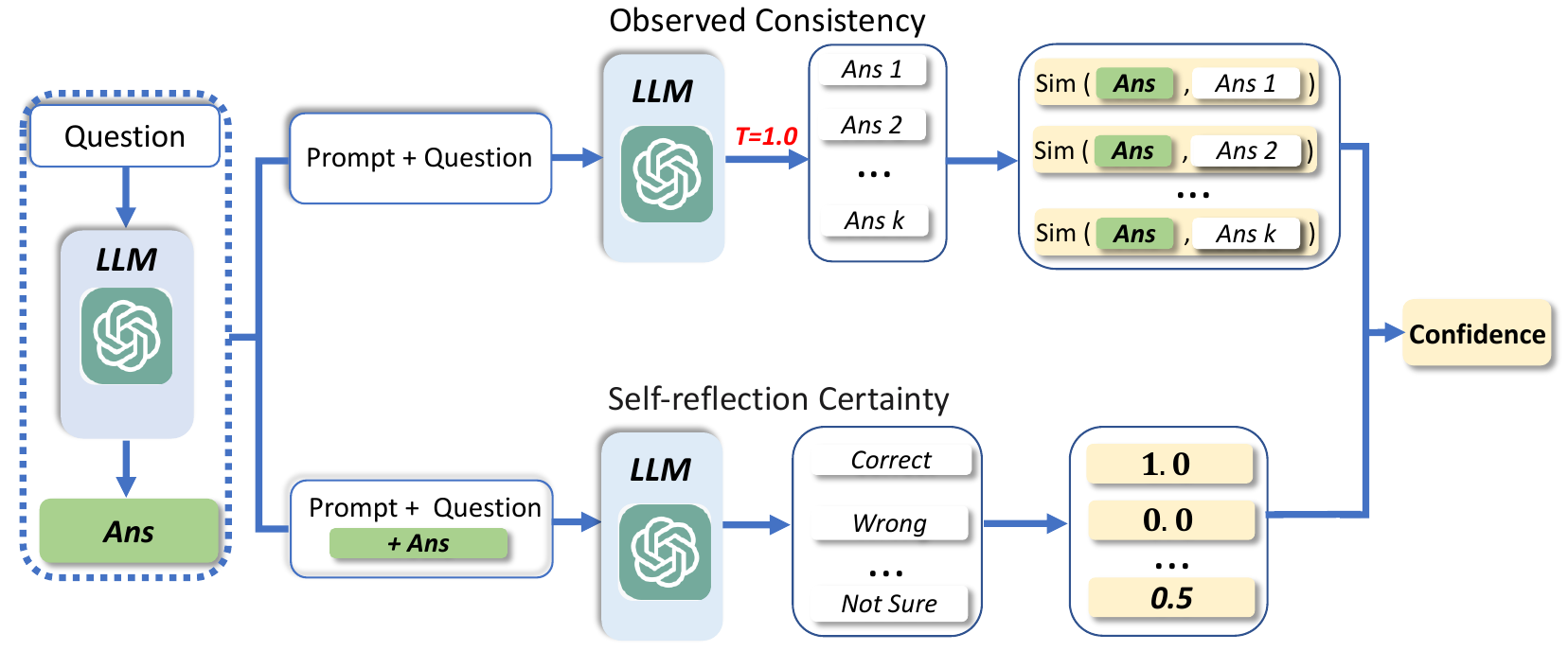}
    \caption{\label{fig:main_task}Pipeline of \modelnameA{}, which can be applied to any LLM API. ($T=1.0$ means temperature sampling with parameter 1.0, Sim ($\cdot$,$\cdot$) means the semantic similarities between two sentences.)}
     \end{subfigure}
     \begin{subfigure}[b]{1.0\textwidth}
         \centering
         \includegraphics[width=\textwidth]{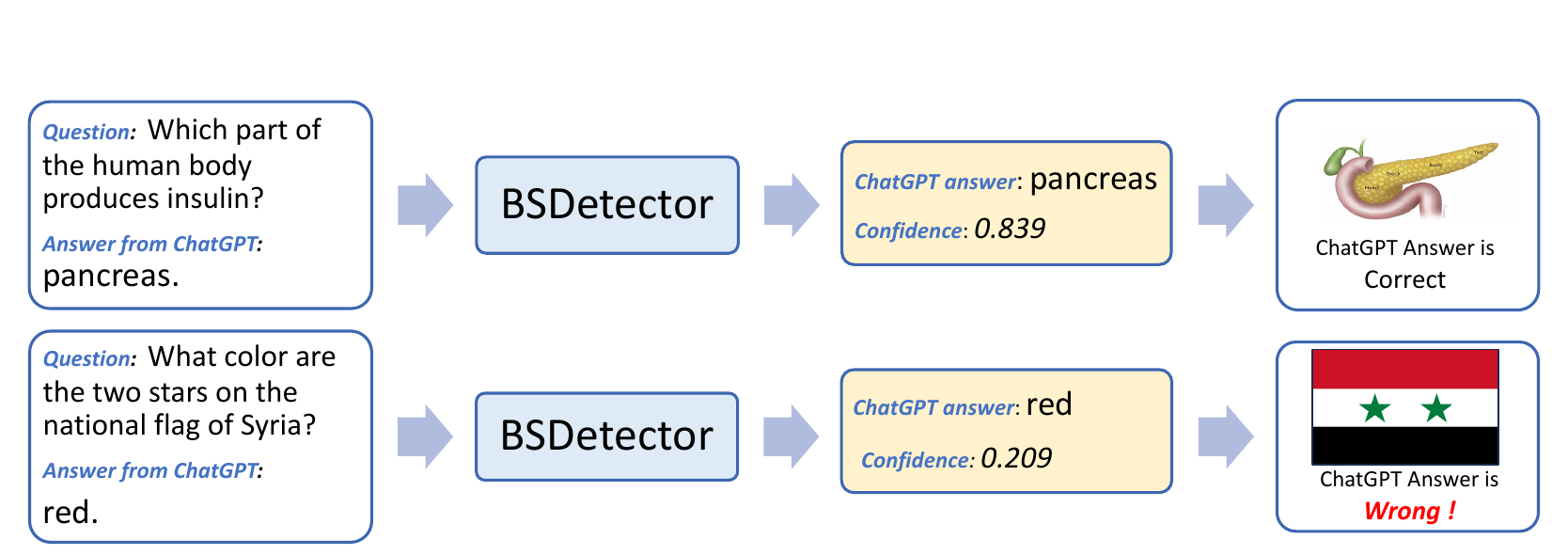}
         \vspace{-1mm}
    \caption{\label{fig:main_example}Two prompts from a Trivia Q\&A dataset \citep{joshi2017triviaqa} and the responses from ChatGPT, along with the associated confidence scores from \modelnameA{}.}
     \end{subfigure}
        \caption{Overview of our LLM uncertainty quantification technique.}
        \label{fig:overview}
\end{figure}

\section{Introduction}

While the promise of Large Language Models (LLMs) and Agents (powered by LLMs) has become evident, their usage in high-value applications remains limited by their \emph{unreliability}. Accessed via black-box APIs (via providers like OpenAI/Anthropic), today's best LLMs have been trained to produce convincing-looking responses and thus often appear overconfident \citep{ji2023survey}. For many input prompts encountered in the wild, the model cannot be certain about the desired response (perhaps because the prompt is vague or is related to a specific fact/event absent from the training dataset), yet these models output plausible-sounding yet wildly incorrect answers in such scenarios. This \emph{hallucination} problem has also plagued traditional supervised learning systems, where it is traditionally addressed via \emph{uncertainty estimation} to know when one can trust a model's prediction \citep{gal2016dropout, lakshminarayanan2017simple, guo2017calibration, liang2017enhancing, fortunato2017bayesian, gal2016theoretically, kuleshov2018accurate}.

In traditional supervised learning, one has access to the training data of the model and its probabilistic estimates, as well as being able to modify the training procedure to improve model  calibration \citep{gal2016dropout, fortunato2017bayesian}. Other traditional uncertainty estimation procedures require the existence of a validation set that can be used for calibration \citep{angelopoulos2021gentle}.  None of this is available for today's best LLMs, which may be given any imaginable prompt rather than (input, output) pairs stemming from a limited distribution. Thus approaches to uncertainty estimation for black-box LLMs must wrap the inference procedure.

Our proposed LLM uncertainty quantification technique, \modelnameA{}, calls the LLM API multiple times with varying prompts and sampling \emph{temperature} values (see Figure \ref{fig:overview}). We expend extra computation in order to quantify how trustworthy the original LLM response is, a worthwhile tradeoff for high-stakes applications. 
Our method is conceptually straightforward, generally applicable across LLM providers (as well as Agent frameworks \citep{langchain} or any  stochastic text $\rightarrow$ text mapping), and produces confidence scores whose values are reliably lower for responses from the LLM that are more likely bad. 

\modelnameA{} confidence scores allow LLMs to be more safely used in high-stakes applications, since we can know which LLM outputs are not to be trusted. Depending on the application, we can adaptively ask a human for an alternative response when the confidence score is low, automatically route the prompt to an alternative LLM provider, or simply respond ``\emph{I don't know}'' when a confident response cannot be generated.  Our experiments reveal that for Question-Answering applications, we can automatically generate \emph{more accurate} answers by sampling multiple responses from the same LLM and selecting the response whose \modelnameA{} confidence estimate is the highest. 

This paper primarily focuses on \emph{Question-Answering} applications, but our same uncertainty estimates can also be applied to estimate how confident the LLM is in its response to a more general prompt. Intuitively, we'd like to see a low confidence score when the LLM outputs: a factually incorrect response to a question, a inaccurate summary requested for a document, or a generated article/message that semantically differs from the intention of the original request. Ensuring this is challenging without control over LLM training, but we can hope that in each of these three scenarios where the model generated a bad response, a well-trained LLM was also likely to output alternative responses (which more closely reflect the desired response). \modelnameA{} is baseed on this intuition, and is observed to produce effective uncertainty estimates with today's top LLMs from OpenAI across prompts from closed and open domain benchmark datasets.

\section{Related Work}

\paragraph{Uncertainty Estimation in Supervised Learning.}

Understanding and quantifying the uncertainties associated with model predictions in traditional supervised learning has a rich history \citep{angelopoulos2021gentle}. Even when trained in a supervised manner, neural network models pose a unique set of challenges for measuring uncertainty and improving calibration \citep{Papadopoulos01, deepbandit}. Much of this work stems from the field of computer vision, where distinct frameworks have been proposed by: \cite{blundell2015weight, gal2016dropout} to approximate Bayesian inference, \cite{lakshminarayanan2017simple, jain2020maximizing} to rely on straightforward deep ensembles, \cite{liang2017enhancing, papernot2018deep} to detect Out-of-Distribution training samples. Parallel ideas for uncertainty estimation with supervised neural works have been developed in natural language processing \citep{fortunato2017bayesian, gal2016theoretically, kuleshov2018accurate}. However, these techniques are not directly applicable to today's best LLMs which are behind black-box APIs with unknown training data.

\paragraph{Uncertainty Estimation for LLMs.}

For estimating the confidence levels tied to responses output by large language models,  \cite{kuhn2023semantic} introduce \emph{semantic entropy}, incorporating linguistic invariances created by shared meanings. However their approach requires access to token-level probabilities from the LLM, which is often not accessible with today's black-box APIs. \cite{kadavath2022language} prompt the models to self-evaluate their answers and directly ask  the LLM to produce the likelihood $P($Answer is True$)$ -- also fine-tuning the model to output better values for its stated likelihood. Relatedly, \cite{lin2022teaching} prompt LLMs to generate both an answer and a level of confidence. \cite{manakul2023selfcheckgpt} propose  a sampling-based approach to detect hallucinated facts. All of these aforementioned approaches train additional models via supervised learning, unlike \modelnameA{} which does not employ any additional training. More recently, \cite{tian2023just} conduct evaluations of computationally feasible methods to extract confidence scores from the probabilities output by LLMs trained via Reinforcement Learning with Human Feedback. \cite{lin2023generating} differentiate between \emph{uncertainty} and \emph{confidence} estimation for LLMs (under their terms, our work is focused on the latter, but without requiring access to the auto-regressive token probability estimates their method is based on). 
The works of \cite{tian2023just} and \cite{lin2023generating} only study limited tasks, and it remains unclear whether their conclusions still hold in the context of reasoning or arithmetic. Here we demonstrate that our method produces effective uncertainty estimates across multiple domains involving reasoning, arithmetic, and knowledge of facts.

\section{\modelnameA{} uncertainty estimation}

When posing a \emph{question} to LLMs, we aim to to estimate how confident we should be that a particular LLM \emph{answer} is correct (or simply ``good'' for more general LLM responses). Specifically, for input question $\bx$, we want to not only obtain an answer $\by$ from the LLM, but also an associated confidence score for this answer $C(x,y)$.
Our confidence assessment derives from two factors: \textbf{Observed Consistency} and \textbf{Self-reflection Certainty}, which respectively are extrinsic and intrinsic evaluations of LLM confidence. Since a well-trained LLM should consider multiple different answers when asked an under-specified question or about something not contained in its training data, Observed Consistency extrinsically measures whether the LLM finds multiple contradictory answers likely to be good responses. Since effective LLMs can reasonably evaluate text from arbitrary agents, Self-Reflection Certainty directly asks the LLM to intrinsically reflect on whether its own previously-generated answer seems correct and how confident it is about this.

\subsection{Observed Consistency}

The first critical measure of model uncertainty is contradiction score amongst possible answers LLMs gives to a particular input questions. Observed Consistency is an extrinsic confidence assessment performed by a user who engages in repeated interactions with LLMs. If a model exhibits strong observed consistency, it's less likely to present alternative responses that are substantially different from its initial answer. The idea was initially inspired by \emph{Self-Consistency} \citep{wang2022self}. While Self-Consistency enhances LLM accuracy in closed-form tasks like arithmetic or commonsense reasoning, it falls short when applied to open-form tasks. Within the Self-consistency approach, an indicator function is used to measure the similarity amongst various likely responses. Here we extend the indicator function to a particular form of semantic similarity based on contradiction ratings, enabling our approach to be used in both open and closed form tasks.

\paragraph{Producing Diverse Output.} Our first action runs the LLM multiple times to produce multiple varied responses. Besides increasing the temperature values (which can only be done so much without getting nonsensical outputs), we can alternatively modify the prompt itself when sampling each response to get a more diverse set of responses for computing the observed consistency.
Here we add a Chain-of-Thoughts (CoT, \cite{wei2022chain}) modification, along with other guidelines for output formatting, to the prompt used to sample these outputs.  
The specific prompt template is illustrated in Figure~\ref{fig:prompt_1}, the outputs produced by this prompt are denoted as $\{\by_1, \by_2, ..., \by_k\}$, where $k$ is the number of sampled outputs. Higher values of $k$ lead to better uncertainty estimates, but require more computation (we found $k=5$ works well enough in practice). 

Note here we only modify the prompt used to sample varied responses for computing the observed consistency, \emph{not} the prompt originally given to produce the original reference response.
We tried alternative prompt modification techniques to encourage greater output diversity (such as adding additional made-up context in the prompt, or encouraging the LLM to answer as a specific persona), but found the CoT modification to work best (Table \ref{tab:ablation_cot}).

\paragraph{Measuring Similarity between Sampled and Original Answer.} After receiving multiple outputs, the following step is to measure the similarities between each element in $\{\by_1, \by_2, ..., \by_k\}$ and original answer $\by$. Instead of using the indicator function to  precisely match two numeric responses (e.g., 1.0 v.s. 2.0) or two choices (e.g.\ A v.s. B), we consider semantic similarities. Not just overall similarities (e.g.\ via LLM embeddings) which are sensitive to variation that does not necessarily indicate the LLM is uncertain, but rather measuring whether the semantics of the two outputs contradict one another or not. A common strategy to estimate this is to use a natural language inference classification system (NLI) \citep{kuhn2023semantic}, which classifies a pair of two text statements $\by_i$ and $\by$ as one of: \emph{entailment}, \emph{neutral}, or \emph{contradiction}. Specifically, the input of NLI is formed by concatenating $\by_i$ and $\by$, and then NLI returns the probabilities $p$ for each of these 3 classes. For each element in $\{\by_1, \by_2, ..., \by_k\}$, we can get the similarity scores with respect to the original reference answer $\by$, denoted as $\{s_1, s_2, ..., s_k\}$.


Note that today's best NLI models \citep{he2020deberta} are significantly smaller than LLMs, and thus the NLI computation to obtain $s_i$ is negligible compared to sampling each LLM answer $\by_i$. However, even the best NLI models were trained on a limited dataset and thus do not always generalize reliably to arbitrary pairs of statements. In particular, we note the contradiction probabilities can be unreliable for single-word statements as encountered in certain closed-form tasks whose answers are likely not well-represented in the original NLI training dataset. To account for this, we additionally incorporate the indicator function in our similarity measure to enhance its stability for closed-form tasks. The indicator function is denoted as 
$r_i = \mathbbm{1}[\by = \by_i] \ \text{ for } i=1,2,..., k.
$

For each element $\by_i$ in $\{\by_1, \by_2, ..., \by_k\}$, we derive the similarity score as: 
\begin{equation}
    o_i = \alpha s_i+ (1-\alpha) r_i 
\end{equation}
Here $0 \le \alpha \le 1$  is a trade-off parameter (fixed at 0.8 in our experiments). It should have larger value the more we trust our NLI model to properly generalize its contradiction estimates. Finally, we average over $k$ samples to obtain the Observed Consistency score for answer $\by$ is $O=\bar o_i$.

\subsection{Self-reflection Certainty}

Our \emph{Self-reflection certainty} is an confidence estimate output by LLM itself when asked follow-up questions encouraging it to directly estimate the correctness of its original answer. Unlike sampling multiple outputs from the model (as in Observed Consistency) or computing likelihoods/entropies based on its token-probabilities which are \emph{extrinsic} operations, self-reflection certainty is an \emph{intrinsic} confidence assessment performed within the LLM. Because today's best LLMs are capable of accounting for rich evidence and evaluation of text \citep{kadavath2022language, lin2022teaching}, such intrinsic assessment via self-reflection 
can reveal additional shortcomings of LLM answers beyond extrinsic consistency assessment. For instance, the LLM might consistently produce the same nonsensical answer to a particular question it is not well equipped to handle, such that the observed consistency score fails to flag this answer as suspicious. Like CoT prompting, self-reflection allows the LLM to employ additional computation to reason more deeply about the correctness of its answer and consider additional evidence it finds relevant. Through these additional steps, the LLM can identify flaws in its original answer, even when it was a high-likelihood (and consistently produced) output for the original prompt.

To specifically calculate self-reflection certainty, we prompt the LLM to state \emph{how confident} it is that its original answer was correct. 
Like \citet{peng2023instruction}, we found asking LLMs to rate their confidence numerically on a continuous scale (0-100) tended to always yield overly high scores (> 90). Instead we ask the LLM to rate its confidence in its original answer via multiple follow-up questions each on a multiple-choice (e.g.\ 3-way) scale. For instance, we instruct the LLM to determine the correctness of the answer by choosing from the options: A) Correct, B) Incorrect, C) I am not sure. Our detailed self-reflection prompt template can be viewed in Figure~\ref{fig:prompt_2}. We assign a numerical score for each choice: A $=1.0$, B $=0.0$ and C $=0.5$, and finally, our self-reported certainty $S$ is the average of these scores over all rounds of such follow-up questions. 

\subsection{Overall Confidence Estimate}
Considering the distinct characteristics of the Observed Consistency and Self-reflection Certainty, we anticipate they might complement each other. \modelnameA{} aggregates the Observed Consistency and Self-reflection Certainty values into an overall confidence score for the LLM response: 
\begin{equation}
    C= \beta O + (1-\beta) S
\end{equation}
Here $0 \le w_2 \le 1$ is a trade-off parameter (fixed as 0.7 in our experiments). It should have larger value the more we trust the LLM's ability to do calibrated self-reflection assessment of arbitrary (question, answer) pairs.

\section{Application: Generating More Reliable Answers from any LLM}
\label{sec:betteranswer}

\begin{figure}[tb]
\centering
\includegraphics[width=1.0\textwidth]{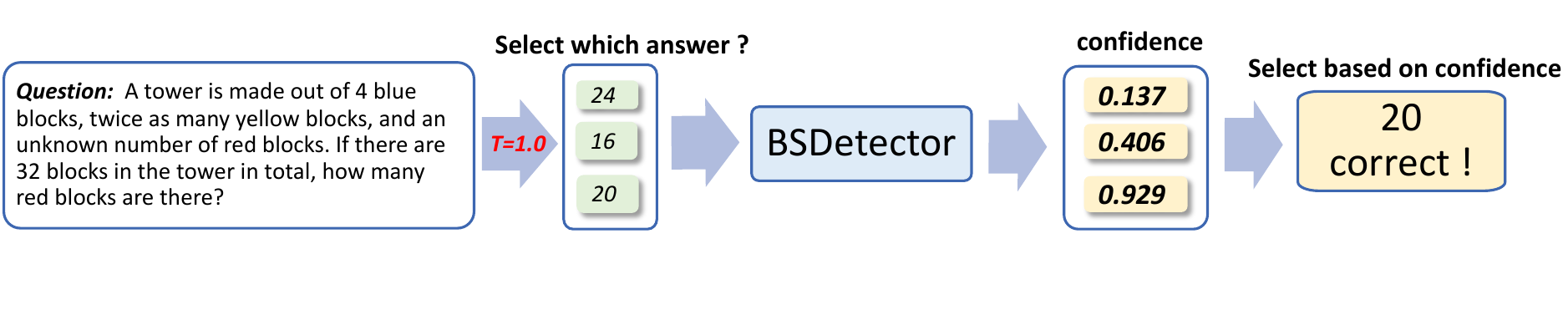}
\vspace{-1em}
\caption{\label{fig:application_1}ChatGPT is used to generate the answers to arithmetic problem "A tower is ..." with temperature sampling $T=1.0$. Subsequently, \modelnameA{} is utilized to select the most confident answer from the three possible answers.}
\end{figure}

One straightforward application of our \modelnameA{} uncertainty estimate is to apply it to (each of) multiple candidate answers produced from the same LLM: $\{\by'_1, \by'_2, ..., \by'_k\}$ (including the original reference answer $\by$ in this set). This assessment allows is to determine which candidate LLM answer $\by'_i$ appears most trustworthy, and return that one instead of always returning $\by$ (see Figure~\ref{fig:application_1}). Specifically, we use the same prompt to ask the LLM to produce several responses via temperature sampling. For each candidate answer, we reuse the same set of previously-described LLM outputs $\{\by_1$, $\by_2$, ..., $\by_k\}$ to compute an observed-consistency score (reducing the computation required to assess the trustworthiness of a set of candidate answers). Following the standard \modelnameA{} procedure, we prompt the LLM to assign a self-reflection certainty to each candidate response. Finally we select the answer with highest \modelnameA{} confidence score, which is often the original reference answer $\by$, but not always. An alternate answer $\by'_i \neq \by$ can be deemed most trustworthy via this procedure only if: the LLM was able to identify fewer likely answers that contradict $\by'_i$ and was more certain about the correctness of $\by'_i$ during the intrinsic self-reflection assessment.

\section{Application:  More reliable LLM-based (automated) evaluation}
\label{sec:llmeval}

In open-domain tasks, it is challenging to evaluate the correctness/quality of answers (irrespective of whether these answers were generated by a LLM or human). Often one resorts to automated evaluation using models like  GPT-3.5-turbo or GPT-4 to assess the correctness
of answers \citep{lin2023generating, chen2023instructzero, alpaca,chen2023takes, xu2023wizardlm,  chen2023you}. Recent instruction fine-tuning techniques such as Alpaca \citep{alpaca} and WizardLM \citep{xu2023wizardlm} also utilize GPT-4 for automated evaluation of generated answers. Even when they are based on advanced LLMs like GPT-4, there remain  
\textbf{questions about the reliability of these LLM-based evaluations}. 

Here we outline two ways to boost the reliability of LLM-based evaluation: \emph{human-in-the--loop} and \emph{fully automated}. Both start by computing BSDetector confidence scores for each LLM-evaluation (these scores estimate not the trustworthiness of the generator of the answers, but rather the evaluator of their correctness). Let $\mathcal{A}$ denote the subset of answers where the corresponding LLM-evaluation had the lowest BSDetector 
 confidence scores (indicating the automated evaluation for this answer is untrustworthy). The gold-standard for evaluating open-domain answers is human inspection, but this is costly. Under a limited labor budget, we can boost the reliability of LLM-based evaluation by having humans only inspect and provide evaluations for the answers in $\mathcal{A}$. In settings where this \emph{human-in-the-loop} approach is not possible, an alternative \emph{fully-automated} way to boost the reliability of LLM-evaluation is to simply omit the answers in $\mathcal{A}$ entirely from the evaluation-set.

\section{Experiments}

\subsection{Calibration of uncertainty estimates}
\paragraph{Datasets.} Our experiments consider numerous question-answering benchmarks listed below. For each example in each benchmark dataset, the true answer is known enabling us to precisely assess the accuracy of LLM responses. We study performance in: GSM8K \citep{cobbe2021training} and SVAMP \citep{patel2021nlp}, datasets composed of grade school math word problems, Commonsense Question Answering (CSQA) \citep{talmor2018commonsenseqa}, a dataset requiring some level of reasoning, and TriviaQA \citep{joshi2017triviaqa}, an open-form trivia question dataset that gauges models' factual knowledge. Because TriviaQA is open-domain, the correct answers provided do not entail all valid solutions, so we also manually validated the accuracy of LLM-generated responses.

\paragraph{Experiment details.} We experiment on two LLMs from OpenAI: Text-Davinci-003 and GPT-3.5 Turbo. The reference answer $\by$ is always produced with the temperature set at 0. To evaluate the confidence of $\by$, we use prompt in Figure~\ref{fig:prompt_1} to generate $k=5$ outputs (unless otherwise stated) with the temperature set at 1.0 (the highest value allowed by the OpenAI API), combined with the indicator function to compute the observed-consistency score. For self-reflection certainty, two follow-up questions in Figure~\ref{fig:prompt_2} are used to assess the correctness of the answer $\by$. As previously described, we combine the observed-consistency and self-reflection certainty to derive the final confidence score.

\paragraph{Evaluation metrics.} Following  \citet{kuhn2023semantic}, we use Area Under the Receiver Operator Characteristic Curve (AUROC) to evaluate the quality of our uncertainty estimates. AUROC represents the likelihood that a correct answer selected at random will have a higher uncertainty score compared to an randomly chosen incorrect answer. A higher AUROC value is preferable, with an ideal AUROC rating being 1, whereas a random uncertainty estimate would yield AUROC $= 0.5$. To evaluate  \emph{generation quality} from the method to get better LLM answers in Section \ref{sec:betteranswer}, we simply rely on the accuracy of LLM answers.

\paragraph{Baseline Methods.} Our study also evaluates the following baseline uncertainty estimation methods: \emph{Likelihood Based Uncertainty} calculates the joint log-probability of a sequence from the autoregressive estimator and normalizes it by the sequence length \citep{malinin2020uncertainty}. While it represents the typical way to estimate \emph{aleatoric} uncertainty in traditional supervised learning and structured prediction \cite{hendrycks2016baseline}, this approach can only can be applied to Text-Davinci-003, since the GPT-3.5 Turbo API does not provide access to token-level probabilities from the model. \emph{Self-reflection Certainty} and \modelnameA{} are introduced in Fig~\ref{fig:main_task}. \emph{Temperature sampling} is equivalent to \modelnameA{} without:  CoT prompting, self-reflection certainty, and the indicator function term inside of the text-similarity metric. 


\paragraph{Results.} Table \ref{tab:main-table} presents the performance results for our various benchmark tasks and uncertainty estimation methods. Here \modelnameA{} significantly outperforms all baselines across datasets, revealing that confidence from \modelnameA{} well aligns with accuracy.

\begin{table}[ht]
\caption{AUROC achieved by different confidence scoring methods across various datasets. }
\label{tab:main-table}
\centering
\begin{tabular}{p{2.35cm}||c|c|c|c|c}
\toprule
\multirow{2}{*}{LLM} & \multirow{2}{*}{Dataset} & Likelihood Based & Temperature & Self-reflection &\multirow{2}{*}{\modelnameA{}} \\
&  &Uncertainty & Sampling & Certainty &  \\
\midrule
\multirow{4}{*}{Text-Davinci-003} 
& GSM8K & 0.647 & 0.614 & 0.521 & \bf 0.867  \\
& CSQA & 0.490 & 0.540 & 0.539 & \bf 0.743 \\
& SVAMP & 0.668  & 0.653 & 0.619 & \bf 0.936\\
& TriviaQA & 0.708 &  0.769 & 0.653 & \bf 0.828 \\ 
\midrule
\multirow{4}{*}{GPT-3.5 Turbo} & 
GSM8K & - & 0.660 & 0.831 & \bf 0.951  \\
& CSQA  & - & 0.583 & 0.506 & \bf 0.769  \\
& SVAMP & - & 0.671 & 0.839 & \bf 0.927 \\
& TriviaQA & - & 0.689 & 0.655 & \bf 0.817 \\
\bottomrule
\end{tabular}
\end{table}

\subsection{Generating More Reliable Answers from any LLM}

In Table \ref{tab:application_1}, we select the response with the highest confidence out of 5 generated responses as described in Section \ref{sec:betteranswer}. For all tasks, \modelnameA{} can identify less accurate responses and notably improve LLM accuracy. Table \ref{tab:application_1} compares this approach against the original single answer $\by$  generated by the LLM (with temperature set to 0), referred to as the \emph{Reference Answer}. While answers produced via the \modelnameA{} filtering procedure from Section  \ref{sec:betteranswer} require 10x as much LLM-inference computation as the Reference Answer, the consistent accuracy gain observed in Table \ref{tab:application_1} makes this worthwhile for high-stakes applications.

\begin{table}[ht]
\caption{Generating more reliable LLM answers. We show the accuracy of each set of answers for the dataset produced from the LLM with a particular method.}
\label{tab:application_1}
\centering
\begin{tabular}{p{2.35cm}||c|c|c}
\toprule
LLM & Dataset & Reference Answer (\%)  & \modelnameA{} (\%) \\
\midrule
\multirow{4}{*}{Text-Davinci-003} & 
GSM8K & 12.50 &  \bf 16.83  \\
& CSQA  & 71.50  & \bf 72.83  \\
& SVAMP & 65.67   & \bf 70.00 \\
& TriviaQA & 69.80 &  \bf 70.50 \\
\midrule
\multirow{4}{*}{GPT-3.5 Turbo} & 
GSM8K & 47.47 & \bf 69.44  \\
& CSQA  & 72.72 & \bf 73.22  \\
& SVAMP & 75.30 & \bf 82.00 \\
& TriviaQA & 73.50 &  \bf 76.00 \\
\bottomrule
\end{tabular}
\end{table}


\subsection{More reliable LLM-based (automated) evaluation}

\begin{figure}
     \centering
     \begin{subfigure}[b]{0.49\textwidth}
         \centering
    \includegraphics[width=\textwidth]{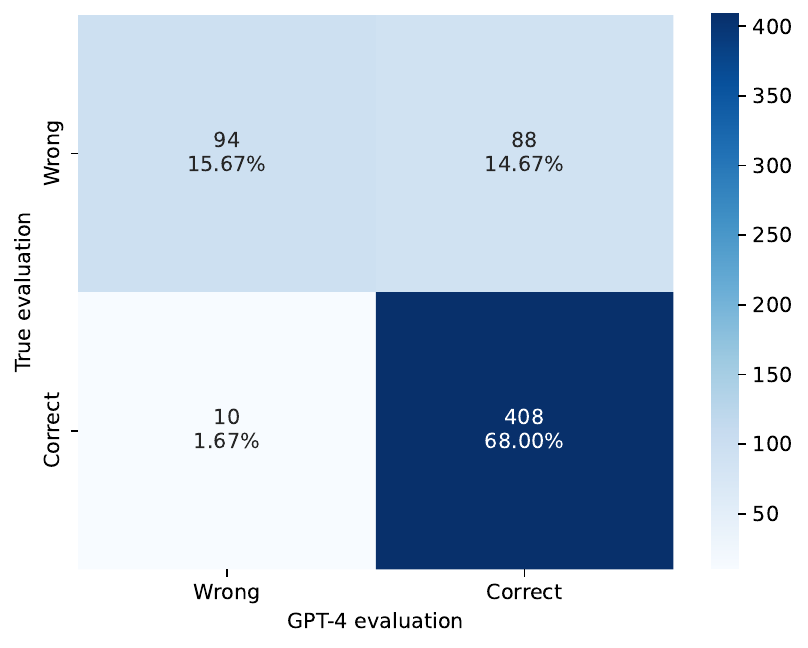}
    \caption{TriviaQA: Overall GPT-4 accuracy: 83.67\%}
        \label{fig:triviaqa_confusion}
     \end{subfigure}
     \hfill
     \begin{subfigure}[b]{0.49\textwidth}
         \centering
    \includegraphics[width=\textwidth]{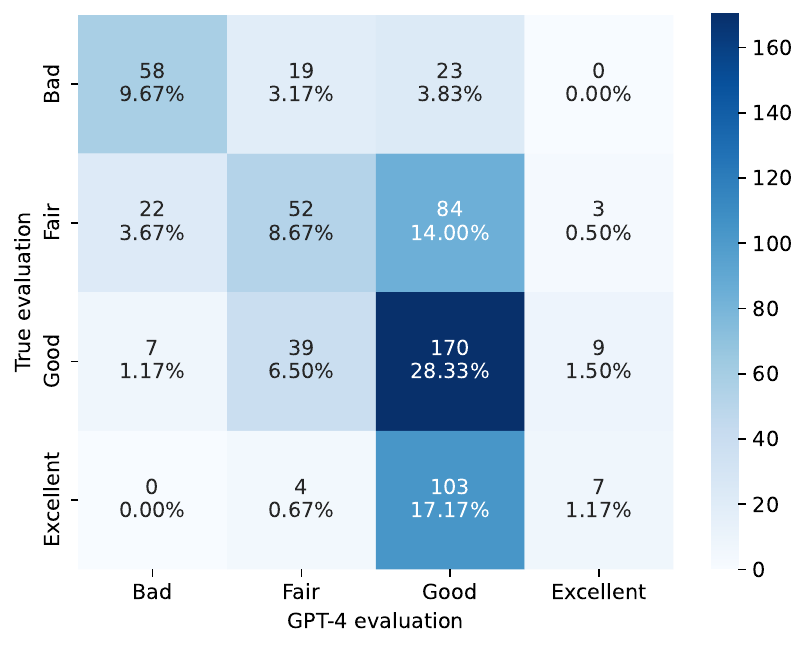}
    \caption{Summarize-from-feedback: GPT-4 MSE: 0.707}
        \label{fig:summary_confusion}
     \end{subfigure}
    \caption{Confusion matrix comparing automated GPT-4 evaluations vs. human evaluations.}
        \label{fig:confusion}
\end{figure}

We first investigate how reliable GPT-4 based evaluation is in practice. First we employ the Text-Davinci-003 model to produce answers for \textbf{TriviaQA} \citep{joshi2017triviaqa}. Subsequently, GPT-4 is given the question and generated answer (from Text-Davinci-003) and asked to designate the answer as correct or incorrect (see the Figure~\ref{fig:prompt_3} for the specific evaluation prompt). Since ground-truth answers are available for TriviaQA, we can report the accuracy of GPT-4 based evaluation, which is only 83.67\% in this setting (Figure \ref{fig:triviaqa_confusion}). Next, we try using GPT-4 to assess the quality of answers. For example, alpaca-eval \citep{alpaca-eval} utilizes GPT-4 to discern which answer from two LLMs is superior but it is unknown how reliable GPT-4 judgements are in their application. To investigate this, we consider a similar task: \textbf{Summarize-from-feedback} \citep{stiennon2020learning}. This dataset provides the original context, a summary derived from that context, and a human assessment of the summary's quality (which we hold out only for reporting purposes here). We employ GPT-4 based evaluation to automatically rate each summary's quality, asking the LLM-evaluator to select from options: Bad, Fair, Good, or Excellent (see the Figure~\ref{fig:prompt_4} for the specific evaluation prompt). Translating these ratings to a 1-4 numerical scale, we report the mean square error (MSE) between these automated GPT-4 ratings vs. the ground truth human ratings. Figure \ref{fig:summary_confusion} shows this  MSE is approximately 0.707.
In both experiments, automated evaluation based on GPT-4 is not as reliable as one would hope to reach trustworthy conclusions.

\begin{figure}
     \centering
     \begin{subfigure}[b]{0.49\textwidth}
         \centering
    \includegraphics[width=\textwidth]{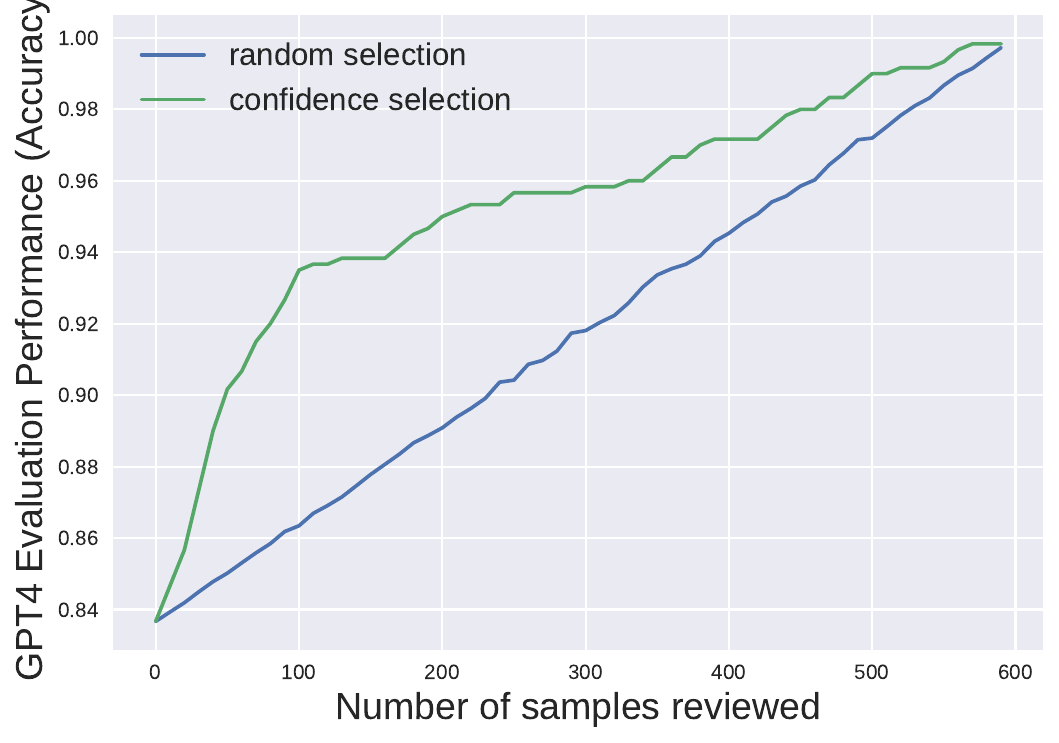}
    \caption{TriviaQA}
        \label{fig:triviaqa_with_human}
     \end{subfigure}
     \hfill
     \begin{subfigure}[b]{0.49\textwidth}
         \centering
    \includegraphics[width=\textwidth]{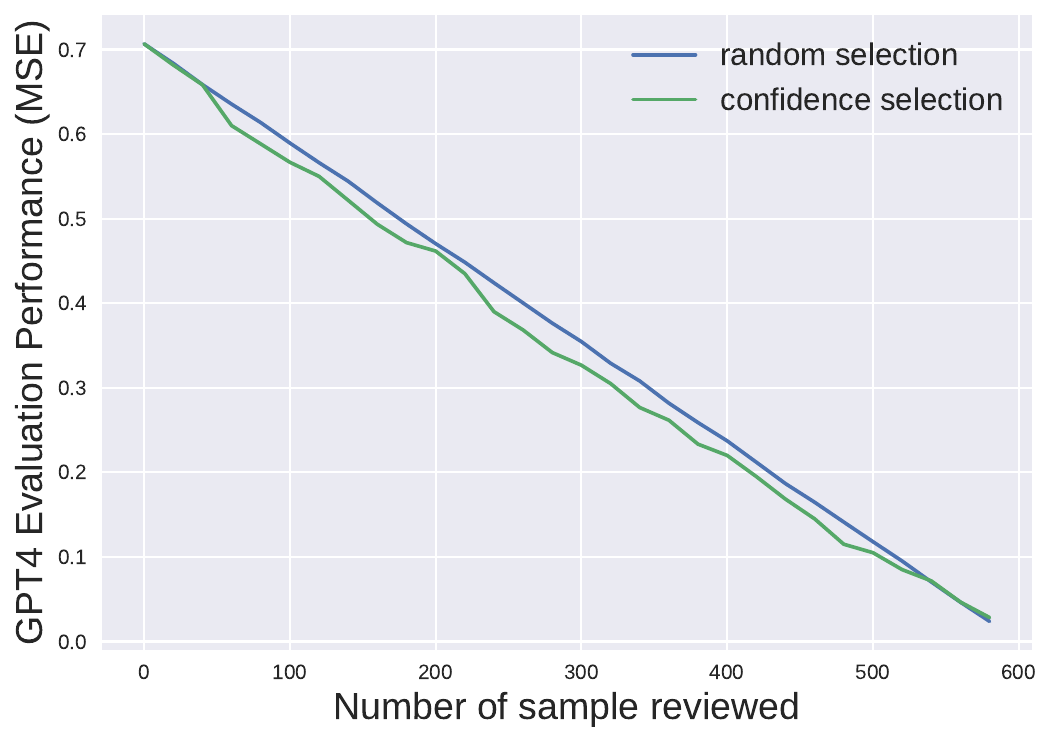}
    \caption{Summarize-from-feedback}
        \label{fig:summary_with_human}
     \end{subfigure}
    \caption{Human in the loop LLM-based evaluation, with the number of answers evaluated by humans varied along the x-axis (remaining answers are auto-evaluated by GPT-4). The resulting accuracy/MSE of the combined set of human + GPT-4 evaluations is shown along y-axis, under confidence-based vs. random selection to decide which subset of answers receive human evaluation.}
        \label{fig:with_human_in_the_loop}
\end{figure}

Finally we study whether \modelnameA{} can help us achieve more reliable evaluations with GPT-4, as described in Section \ref{sec:llmeval}. We consider the TriviaQA and Summarize-from-feedback datasets with the same GPT-4 model and evaluation prompts from the previous paragraph, and compute \modelnameA{} confidence scores for the GPT-4 evaluator as described in Section \ref{sec:llmeval}. We first consider the \emph{human-in-the-loop} setting, where a human provides the evaluation for answers in $\mathcal{A}$, defined as the subset of answers where the corresponding GPT-4 evaluation has \modelnameA{} confidence score amongst the $K$ lowest values. We compare the resulting set of combined automated + human evaluations (\textbf{confidence selection}) against a baseline set of combined automated + human evaluations, where the subset of answers evaluated by a human is chosen via \textbf{random selection} (rather than based on our confidence score). Figure \ref{fig:with_human_in_the_loop} depicts the performance of the resulting \emph{human-in-the-loop} evaluation vs. the number of answers $K$ evaluated by a human (remaining answers are all auto-evaluated by GPT-4). Across both datasets, guiding the human-the-loop evaluation based on \modelnameA{} confidence yields more reliable evaluations.


To conclude, we study the \emph{fully-automated} approach to LLM-based evaluation from Section \ref{sec:llmeval}, which offers a labor-free way to utilize the \modelnameA{} confidence scores. Recall in this approach we simply omit the subset of answers in $\mathcal{A}$ from the evaluation-set entirely. We can then compute the average evaluation-score from GPT-4 as an overall quality estimate for the collection of generated answers. Intuitively, we do not want to include answers in this average whose GPT-4 evaluation is highly uncertain (to reduce variance), but discarding answers shrinks the remaining evaluation-set thus increasing variance of the resulting average.

\begin{figure}
     \centering
     \begin{subfigure}[b]{0.495\textwidth}
         \centering
         \includegraphics[width=\textwidth]{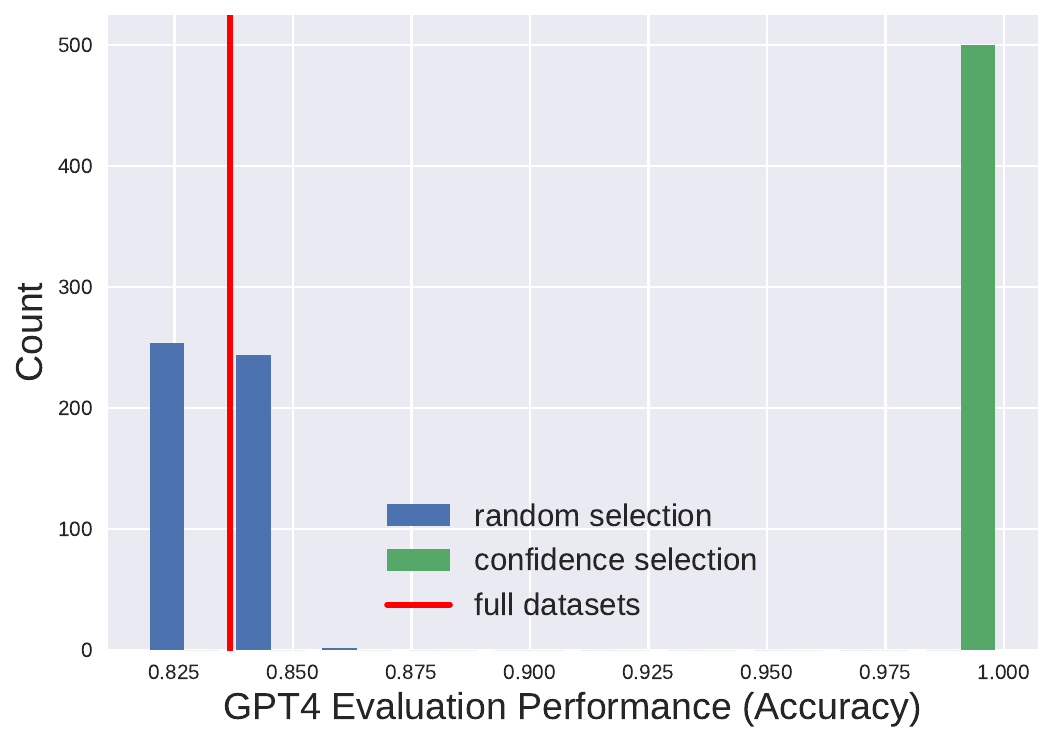}
         \caption{TriviaQA}
        \label{fig:triviaqa_without_human}
     \end{subfigure}
     \hfill
     \begin{subfigure}[b]{0.495\textwidth}
         \centering
         \includegraphics[width=\textwidth]{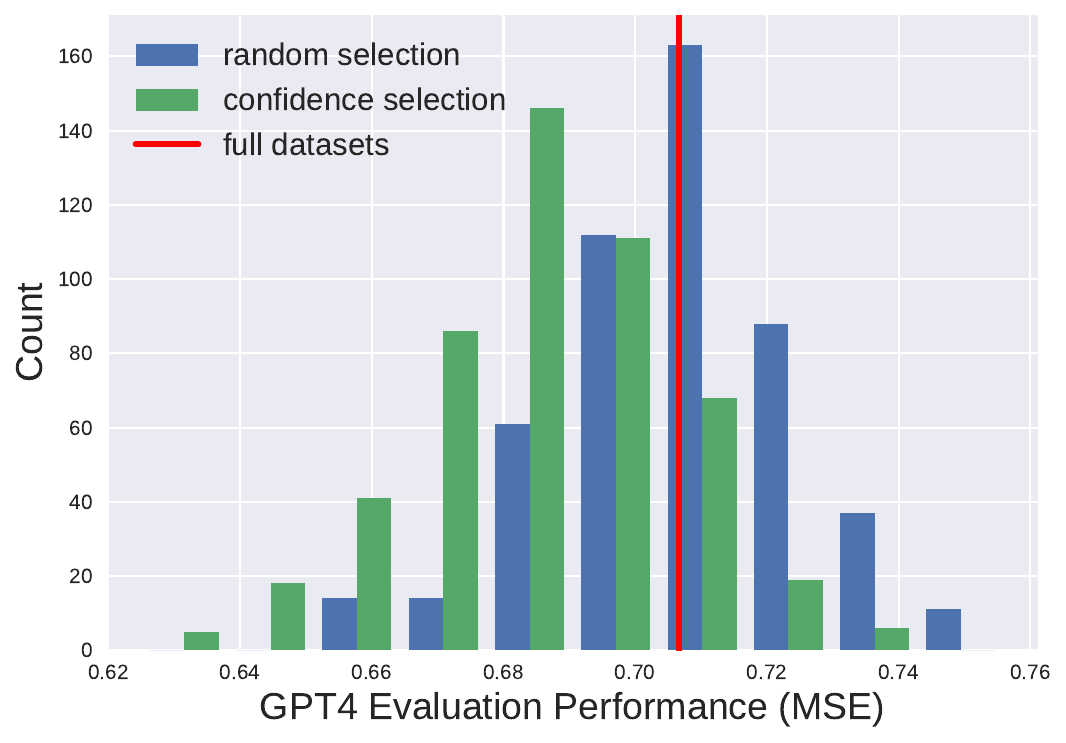}
         \caption{Summarize-from-feedback}
        \label{fig:summary_without_human}
     \end{subfigure}
        \caption{Fully-automated GPT-4 based evaluation, assessing the accuracy/MSE over many replicate datasets (observed counts amongst replicates on y-axis). By discarding the bottom 20\% of evaluations with the lowest confidence, the average GPT-4 evaluation score consistently reaches an accuracy of 1.0 on TriviaQA, indicating completely trustworthy LLM-based evaluations (and the MSE of the average GPT-4 score consistently improves compared to the full dataset or discarding a random 20\%). 
        }
        \label{fig:without_human_in_the_loop}
\end{figure}

Evaluating the impact of these variance changes  requires statistical repetition, so we repeat the following procedure 500 times:
For both datasets (TriviaQA, Summarize-from-feedback), we select 500 answers and calculate the average GPT4 evaluation-score over these answers. We call these the \emph{full} dataset and the resulting average is the baseline score (estimator), whose accuracy/MSE we report against the average human evaluation score across the full dataset (estimand).
To utilize \modelnameA{} for a more reliable estimator of the average human-evaluation score, we simply remove the 20\% of answers with the lowest confidence scores for the corresponding GPT-4 evaluation, and compute the average GPT-4 evaluation score over the remaining 400 answers.
As a sanity check, we also repeat this procedure but this time randomly dropping 20\% of the answers (rather than based on confidence score), which purely increases the variance of resulting average GPT-4 evaluation score with no benefits. Figure \ref{fig:without_human_in_the_loop} shows the resulting deviation between average GPT-evaluation score and average human evaluation score over all of these statistical replicate experiments. Across both datasets, we get more reliable average LLM-evaluation scores by discarding the answers with the lowest confidence scores for the corresponding LLM-evaluation. Preventing the high-uncertainty LLM-evaluations from corrupting the average evaluation score is clearly worth the variance-penalty paid by shrinking the size of the evaluation set.

\section{Discussion}

This paper presents \modelnameA{}, a method designed to identify unreliable or speculative answers from LLMs by computing a confidence score for its generated outputs. Our uncertainty estimates are applicable to any LLM, even those only accessible via a black-box API, and combine both intrinsic and extrinsic evaluations of confidence. By sampling multiple LLM answers and selecting the one with the highest associated confidence score, we can produce more accurate responses from the same LLM without any additional training. 
One open question is how to minimize the computational cost to achieve a desired level of confidence score calibration (for instance via adaptive produces that expend more compute only for assessing those answers whose confidence is hardest to estimate). 
Due to its simplicity and generality, we expect \modelnameA{} uncertainty estimation to find many applications across diverse domains/tasks, beyond the studies in this paper on: deciding what LLM responses cannot be trusted, and enhancing the accuracy of LLM answers and LLM-based evaluation. 

\clearpage

\bibliography{main}

\begin{thebibliography}{38}
\providecommand{\natexlab}[1]{#1}
\providecommand{\url}[1]{\texttt{#1}}
\expandafter\ifx\csname urlstyle\endcsname\relax
  \providecommand{\doi}[1]{doi: #1}\else
  \providecommand{\doi}{doi: \begingroup \urlstyle{rm}\Url}\fi

\bibitem[Angelopoulos \& Bates(2021)Angelopoulos and
  Bates]{angelopoulos2021gentle}
Anastasios~N Angelopoulos and Stephen Bates.
\newblock A gentle introduction to conformal prediction and distribution-free
  uncertainty quantification.
\newblock \emph{arXiv preprint arXiv:2107.07511}, 2021.

\bibitem[Blundell et~al.(2015)Blundell, Cornebise, Kavukcuoglu, and
  Wierstra]{blundell2015weight}
Charles Blundell, Julien Cornebise, Koray Kavukcuoglu, and Daan Wierstra.
\newblock Weight uncertainty in neural networks.
\newblock \emph{arXiv preprint arXiv:1505.05424}, 2015.

\bibitem[Chase(2022)]{langchain}
Harrison Chase.
\newblock {LangChain}, 2022.
\newblock URL \url{https://github.com/hwchase17/langchain}.

\bibitem[Chen et~al.(2023{\natexlab{a}})Chen, Chen, Huang, and
  Zhou]{chen2023you}
Jiuhai Chen, Lichang Chen, Heng Huang, and Tianyi Zhou.
\newblock When do you need chain-of-thought prompting for chatgpt?
\newblock \emph{arXiv preprint arXiv:2304.03262}, 2023{\natexlab{a}}.

\bibitem[Chen et~al.(2023{\natexlab{b}})Chen, Chen, and Zhou]{chen2023takes}
Jiuhai Chen, Lichang Chen, and Tianyi Zhou.
\newblock It takes one to tango but more make trouble? in-context training with
  different number of demonstrations.
\newblock \emph{arXiv preprint arXiv:2303.08119}, 2023{\natexlab{b}}.

\bibitem[Chen et~al.(2023{\natexlab{c}})Chen, Chen, Goldstein, Huang, and
  Zhou]{chen2023instructzero}
Lichang Chen, Jiuhai Chen, Tom Goldstein, Heng Huang, and Tianyi Zhou.
\newblock Instructzero: Efficient instruction optimization for black-box large
  language models.
\newblock \emph{arXiv preprint arXiv:2306.03082}, 2023{\natexlab{c}}.

\bibitem[Cobbe et~al.(2021)Cobbe, Kosaraju, Bavarian, Hilton, Nakano, Hesse,
  and Schulman]{cobbe2021training}
Karl Cobbe, Vineet Kosaraju, Mohammad Bavarian, Jacob Hilton, Reiichiro Nakano,
  Christopher Hesse, and John Schulman.
\newblock Training verifiers to solve math word problems.
\newblock \emph{CoRR}, abs/2110.14168, 2021.

\bibitem[Fortunato et~al.(2017)Fortunato, Blundell, and
  Vinyals]{fortunato2017bayesian}
Meire Fortunato, Charles Blundell, and Oriol Vinyals.
\newblock Bayesian recurrent neural networks.
\newblock \emph{arXiv preprint arXiv:1704.02798}, 2017.

\bibitem[Gal \& Ghahramani(2016{\natexlab{a}})Gal and
  Ghahramani]{gal2016dropout}
Yarin Gal and Zoubin Ghahramani.
\newblock Dropout as a bayesian approximation: Representing model uncertainty
  in deep learning.
\newblock In \emph{international conference on machine learning}, pp.\
  1050--1059. PMLR, 2016{\natexlab{a}}.

\bibitem[Gal \& Ghahramani(2016{\natexlab{b}})Gal and
  Ghahramani]{gal2016theoretically}
Yarin Gal and Zoubin Ghahramani.
\newblock A theoretically grounded application of dropout in recurrent neural
  networks.
\newblock \emph{Advances in neural information processing systems}, 29,
  2016{\natexlab{b}}.

\bibitem[Guo et~al.(2017)Guo, Pleiss, Sun, and Weinberger]{guo2017calibration}
Chuan Guo, Geoff Pleiss, Yu~Sun, and Kilian~Q Weinberger.
\newblock On calibration of modern neural networks.
\newblock In \emph{International conference on machine learning}, pp.\
  1321--1330. PMLR, 2017.

\bibitem[He et~al.(2020)He, Liu, Gao, and Chen]{he2020deberta}
Pengcheng He, Xiaodong Liu, Jianfeng Gao, and Weizhu Chen.
\newblock Deberta: Decoding-enhanced bert with disentangled attention.
\newblock \emph{arXiv preprint arXiv:2006.03654}, 2020.

\bibitem[Hendrycks \& Gimpel(2017)Hendrycks and Gimpel]{hendrycks2016baseline}
Dan Hendrycks and Kevin Gimpel.
\newblock A baseline for detecting misclassified and out-of-distribution
  examples in neural networks.
\newblock In \emph{International Conference on Learning Representations}, 2017.

\bibitem[Jain et~al.(2020)Jain, Liu, Mueller, and Gifford]{jain2020maximizing}
Siddhartha Jain, Ge~Liu, Jonas Mueller, and David Gifford.
\newblock Maximizing overall diversity for improved uncertainty estimates in
  deep ensembles.
\newblock In \emph{Proceedings of the AAAI conference on artificial
  intelligence}, 2020.

\bibitem[Ji et~al.(2023)Ji, Lee, Frieske, Yu, Su, Xu, Ishii, Bang, Madotto, and
  Fung]{ji2023survey}
Ziwei Ji, Nayeon Lee, Rita Frieske, Tiezheng Yu, Dan Su, Yan Xu, Etsuko Ishii,
  Ye~Jin Bang, Andrea Madotto, and Pascale Fung.
\newblock Survey of hallucination in natural language generation.
\newblock \emph{ACM Computing Surveys}, 55\penalty0 (12):\penalty0 1--38, 2023.

\bibitem[Joshi et~al.(2017)Joshi, Choi, Weld, and
  Zettlemoyer]{joshi2017triviaqa}
Mandar Joshi, Eunsol Choi, Daniel~S Weld, and Luke Zettlemoyer.
\newblock Triviaqa: A large scale distantly supervised challenge dataset for
  reading comprehension.
\newblock \emph{arXiv preprint arXiv:1705.03551}, 2017.

\bibitem[Kadavath et~al.(2022)Kadavath, Conerly, Askell, Henighan, Drain,
  Perez, Schiefer, Hatfield-Dodds, DasSarma, Tran-Johnson,
  et~al.]{kadavath2022language}
Saurav Kadavath, Tom Conerly, Amanda Askell, Tom Henighan, Dawn Drain, Ethan
  Perez, Nicholas Schiefer, Zac Hatfield-Dodds, Nova DasSarma, Eli
  Tran-Johnson, et~al.
\newblock Language models (mostly) know what they know.
\newblock \emph{arXiv preprint arXiv:2207.05221}, 2022.

\bibitem[Kuhn et~al.(2023)Kuhn, Gal, and Farquhar]{kuhn2023semantic}
Lorenz Kuhn, Yarin Gal, and Sebastian Farquhar.
\newblock Semantic uncertainty: Linguistic invariances for uncertainty
  estimation in natural language generation.
\newblock \emph{arXiv preprint arXiv:2302.09664}, 2023.

\bibitem[Kuleshov et~al.(2018)Kuleshov, Fenner, and
  Ermon]{kuleshov2018accurate}
Volodymyr Kuleshov, Nathan Fenner, and Stefano Ermon.
\newblock Accurate uncertainties for deep learning using calibrated regression.
\newblock In \emph{International conference on machine learning}, pp.\
  2796--2804. PMLR, 2018.

\bibitem[Lakshminarayanan et~al.(2017)Lakshminarayanan, Pritzel, and
  Blundell]{lakshminarayanan2017simple}
Balaji Lakshminarayanan, Alexander Pritzel, and Charles Blundell.
\newblock Simple and scalable predictive uncertainty estimation using deep
  ensembles.
\newblock \emph{Advances in neural information processing systems}, 30, 2017.

\bibitem[Liang et~al.(2017)Liang, Li, and Srikant]{liang2017enhancing}
Shiyu Liang, Yixuan Li, and Rayadurgam Srikant.
\newblock Enhancing the reliability of out-of-distribution image detection in
  neural networks.
\newblock \emph{arXiv preprint arXiv:1706.02690}, 2017.

\bibitem[Lin et~al.(2022)Lin, Hilton, and Evans]{lin2022teaching}
Stephanie Lin, Jacob Hilton, and Owain Evans.
\newblock Teaching models to express their uncertainty in words.
\newblock \emph{arXiv preprint arXiv:2205.14334}, 2022.

\bibitem[Lin et~al.(2023)Lin, Trivedi, and Sun]{lin2023generating}
Zhen Lin, Shubhendu Trivedi, and Jimeng Sun.
\newblock Generating with confidence: Uncertainty quantification for black-box
  large language models.
\newblock \emph{arXiv preprint arXiv:2305.19187}, 2023.

\bibitem[Malinin \& Gales(2020)Malinin and Gales]{malinin2020uncertainty}
Andrey Malinin and Mark Gales.
\newblock Uncertainty estimation in autoregressive structured prediction.
\newblock \emph{arXiv preprint arXiv:2002.07650}, 2020.

\bibitem[Manakul et~al.(2023)Manakul, Liusie, and
  Gales]{manakul2023selfcheckgpt}
Potsawee Manakul, Adian Liusie, and Mark~JF Gales.
\newblock Selfcheckgpt: Zero-resource black-box hallucination detection for
  generative large language models.
\newblock \emph{arXiv preprint arXiv:2303.08896}, 2023.

\bibitem[Papadopoulos et~al.(2001)Papadopoulos, Edwards, and
  Murray]{Papadopoulos01}
G.~Papadopoulos, P.~J. Edwards, and A.~F. Murray.
\newblock {Confidence estimation methods for neural networks: A practical
  comparison}.
\newblock \emph{IEEE Transactions on Neural Networks}, 12:\penalty0 1278--1287,
  2001.

\bibitem[Papernot \& McDaniel(2018)Papernot and McDaniel]{papernot2018deep}
Nicolas Papernot and Patrick McDaniel.
\newblock Deep k-nearest neighbors: Towards confident, interpretable and robust
  deep learning.
\newblock \emph{arXiv preprint arXiv:1803.04765}, 2018.

\bibitem[Patel et~al.(2021)Patel, Bhattamishra, and Goyal]{patel2021nlp}
Arkil Patel, Satwik Bhattamishra, and Navin Goyal.
\newblock Are {NLP} models really able to solve simple math word problems?
\newblock 2021.

\bibitem[Peng et~al.(2023)Peng, Li, He, Galley, and Gao]{peng2023instruction}
Baolin Peng, Chunyuan Li, Pengcheng He, Michel Galley, and Jianfeng Gao.
\newblock Instruction tuning with gpt-4.
\newblock \emph{arXiv preprint arXiv:2304.03277}, 2023.

\bibitem[Riquelme et~al.(2018)Riquelme, Tucker, and Snoek]{deepbandit}
Carlos Riquelme, George Tucker, and Jasper Snoek.
\newblock Deep bayesian bandits showdown: An empirical comparison of bayesian
  deep networks for thompson sampling.
\newblock In \emph{International Conference on Learning Representations}, 2018.

\bibitem[Stiennon et~al.(2020)Stiennon, Ouyang, Wu, Ziegler, Lowe, Voss,
  Radford, Amodei, and Christiano]{stiennon2020learning}
Nisan Stiennon, Long Ouyang, Jeffrey Wu, Daniel Ziegler, Ryan Lowe, Chelsea
  Voss, Alec Radford, Dario Amodei, and Paul~F Christiano.
\newblock Learning to summarize with human feedback.
\newblock \emph{Advances in Neural Information Processing Systems},
  33:\penalty0 3008--3021, 2020.

\bibitem[Talmor et~al.(2019)Talmor, Herzig, Lourie, and
  Berant]{talmor2018commonsenseqa}
Alon Talmor, Jonathan Herzig, Nicholas Lourie, and Jonathan Berant.
\newblock {CommonsenseQA}: {A} question answering challenge targeting
  commonsense knowledge.
\newblock 2019.

\bibitem[Taori et~al.(2023)Taori, Gulrajani, Zhang, Dubois, Li, Guestrin,
  Liang, and Hashimoto]{alpaca}
Rohan Taori, Ishaan Gulrajani, Tianyi Zhang, Yann Dubois, Xuechen Li, Carlos
  Guestrin, Percy Liang, and Tatsunori~B. Hashimoto.
\newblock Stanford alpaca: An instruction-following llama model.
\newblock \url{https://github.com/tatsu-lab/stanford_alpaca}, 2023.

\bibitem[Tian et~al.(2023)Tian, Mitchell, Zhou, Sharma, Rafailov, Yao, Finn,
  and Manning]{tian2023just}
Katherine Tian, Eric Mitchell, Allan Zhou, Archit Sharma, Rafael Rafailov,
  Huaxiu Yao, Chelsea Finn, and Christopher~D Manning.
\newblock Just ask for calibration: Strategies for eliciting calibrated
  confidence scores from language models fine-tuned with human feedback.
\newblock \emph{arXiv preprint arXiv:2305.14975}, 2023.

\bibitem[Wang et~al.(2022)Wang, Wei, Schuurmans, Le, Chi, Narang, Chowdhery,
  and Zhou]{wang2022self}
Xuezhi Wang, Jason Wei, Dale Schuurmans, Quoc Le, Ed~Chi, Sharan Narang,
  Aakanksha Chowdhery, and Denny Zhou.
\newblock Self-consistency improves chain of thought reasoning in language
  models.
\newblock \emph{arXiv preprint arXiv:2203.11171}, 2022.

\bibitem[Wei et~al.(2022)Wei, Wang, Schuurmans, Bosma, Xia, Chi, Le, Zhou,
  et~al.]{wei2022chain}
Jason Wei, Xuezhi Wang, Dale Schuurmans, Maarten Bosma, Fei Xia, Ed~Chi, Quoc~V
  Le, Denny Zhou, et~al.
\newblock Chain-of-thought prompting elicits reasoning in large language
  models.
\newblock \emph{Advances in Neural Information Processing Systems},
  35:\penalty0 24824--24837, 2022.

\bibitem[Xu et~al.(2023)Xu, Sun, Zheng, Geng, Zhao, Feng, Tao, and
  Jiang]{xu2023wizardlm}
Can Xu, Qingfeng Sun, Kai Zheng, Xiubo Geng, Pu~Zhao, Jiazhan Feng, Chongyang
  Tao, and Daxin Jiang.
\newblock Wizardlm: Empowering large language models to follow complex
  instructions.
\newblock \emph{arXiv preprint arXiv:2304.12244}, 2023.

\bibitem[Yann(2023)]{alpaca-eval}
Rohan Yann.
\newblock {alpaca-eval}, 2023.
\newblock URL \url{https://github.com/tatsu-lab/alpaca_eval}.

\end{thebibliography}
\bibliographystyle{iclr2024_conference}

\clearpage
\appendix
\section{Appendix}

\subsection{Details about NLI model}
Specifically, the input of NLI is formed by concatenating $\by_i$ and $\by$, and then NLI returns the probabilities $p$ for each of these 3 classes. Here we choose $1-p_{contradiction}$ (output by an already trained NLI system \citep{he2020deberta}) as our similarity between two sampled LLM outputs. To mitigate positional bias within the NLI system, we consider both orders $(\by_i, \by)$ and $(\by, \by_i)$, producing  $1-p_{contradiction}$ and $1-p'_{contradiction}$ for each order and averaging these two values into a single similarity score. The similarity scores using NLI to assess each sampled LLM answer for contradictions with respect to the original reference answer are denoted:
\[
s_i = \frac{1}{2}(1-p_{contradiction}+1-p'_{contradiction}) \ \ \text{ for } i=1,2,..., k.
\]

\subsection{Compute costs}
The compute costs associated with various uncertainty methods differ. Uncertainty based on autoregressive likelihood is the most cost-effective, requiring only a single API call that returns the token-level probability. However, this cannot be implemented on GPT-3.5 Turbo since it does not provide token-level probabilities. While \modelnameA{} incurs a slight  additional cost for self-certainty reflection in comparison to the baseline Temperature Sampling approach, Table ~\ref{tab:ablation_num} shows that even when we double the number of outputs from Temperature Sampling (thus allowing it far more compute than our approach), its performance remains inferior to \modelnameA{}.

\subsection{Prompts used in \modelnameA{}}

\begin{figure}[tb]
     \centering
     \begin{subfigure}[b]{1.0\textwidth}
         \centering
         \includegraphics[width=\textwidth]{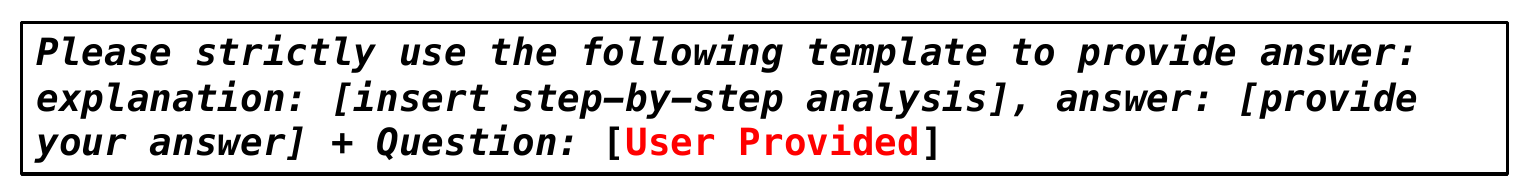}
    \caption{\label{fig:prompt_1}Prompt template for Observed Consistency}
     \end{subfigure}
     \hfill
     \begin{subfigure}[b]{1.0\textwidth}
         \centering
         \includegraphics[width=\textwidth]{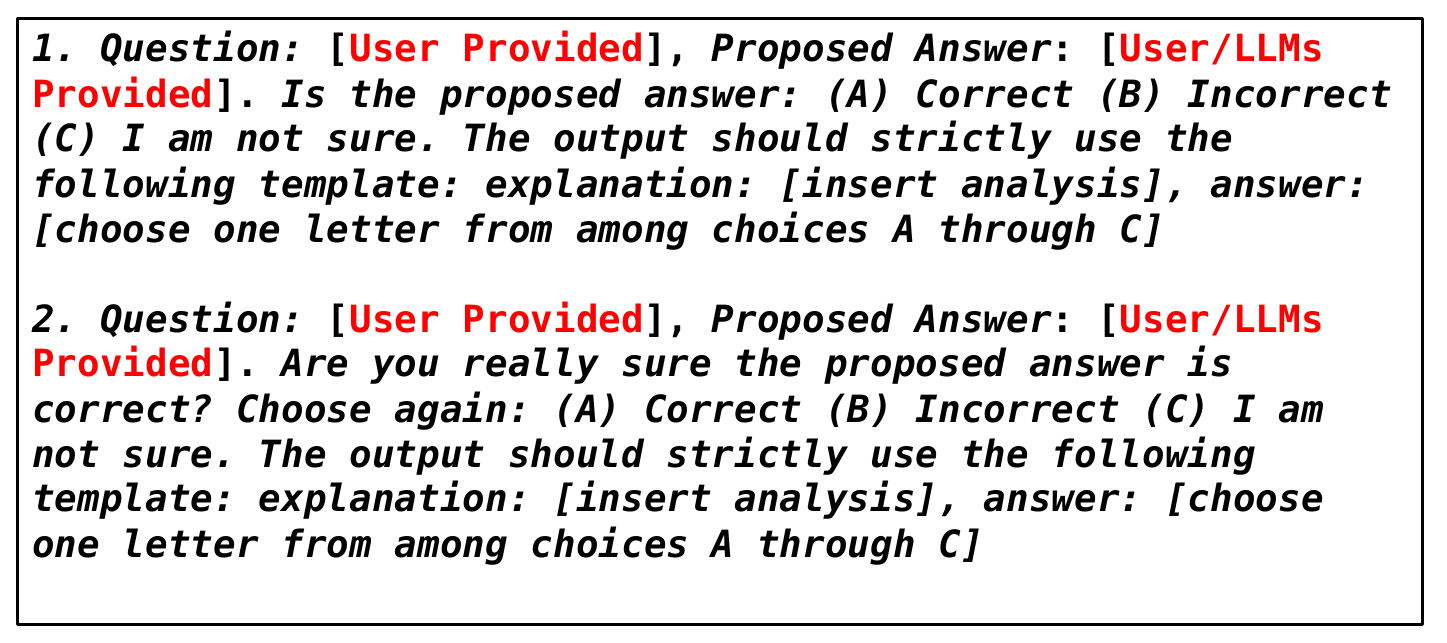}
    \caption{\label{fig:prompt_2}Prompt template for Self-reflection Certainty}
     \end{subfigure}
          \hfill
     \begin{subfigure}[b]{1.0\textwidth}
         \centering
         \includegraphics[width=\textwidth]{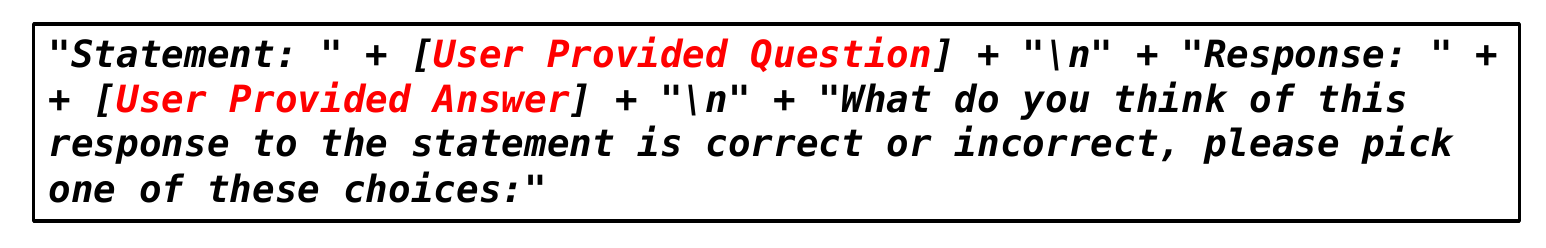}
    \caption{\label{fig:prompt_3}Prompt template for triviaQA  in the application of using \modelnameA{} as an evaluator. }
     \end{subfigure}
          \hfill
     \begin{subfigure}[b]{1.0\textwidth}
         \centering
         \includegraphics[width=\textwidth]{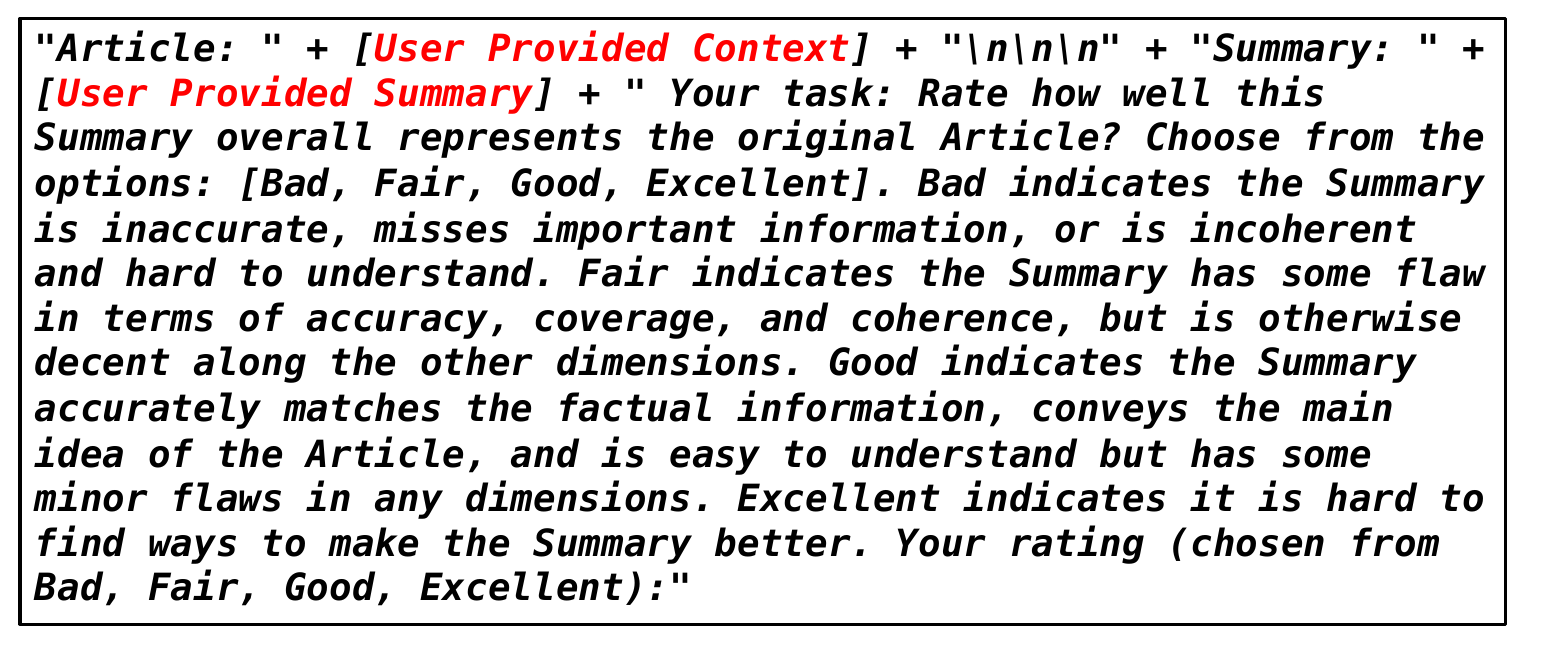}
    \caption{\label{fig:prompt_4}Prompt template for Summarize-from-feedback in the application of using \modelnameA{} as an evaluator. }
     \end{subfigure}
        \caption{\label{fig:prompt}Prompts used to produce the confidence score in \modelnameA{}.}
\end{figure}

Figure \ref{fig:prompt} show the prompts used in \modelnameA{}.

\subsection{Ablation Study}

In this section, we study that whether each component is required to achieve high quality. Our investigation leads to the following primary insights: 1) Enhancing the number of outputs and integrating CoT prompt in Observed Consistency result in a greater variety of responses, thereby making the confidence estimation more reliable. 2) Our similarity metric is crucial for capturing the variation between different responses. 
\subsubsection{Increasing the number of outputs and integrating CoT prompt introduce more diversity?}
Table \ref{tab:ablation_num} shows an ablation study involving the number of outputs in  Observed Consistency, we compare 5 and 10 outputs, observing that for each dataset 10 outputs outperforms 5 outputs. However, for GSM8K, SVAMP, and TriviaQA, the gain from 5 to 10 outputs is marginal. Given the trade-off between cost and performance, and considering that doubling the API calls results in only a slight improvement, we decide to stick with 5 outputs in our experiments. Table \ref{tab:ablation_cot} indicates that CoT is essential for introducing the diversity of responses and achieving the good confidence estimation performance.

\begin{table}[ht]
    \caption{Ablation study}
    \begin{subtable}{.45\linewidth}
      \centering
      \captionsetup{width=.8\linewidth}
        \caption{\label{tab:ablation_num}AUC of \modelnameA{} with different numbers of outputs.}
        \centering
        \begin{tabular}{p{1.2cm}||c|c}
        \toprule &
        5 outputs & 10 outputs\\
        \midrule
        GSM8K  & 0.951 & 0.961 \\
        CSQA   & 0.769  & 0.802  \\
        SVAMP  & 0.927 & 0.937 \\
        TriviaQA  & 0.817 & 0.814 \\
        \bottomrule
        \end{tabular}
    \end{subtable}%
    \begin{subtable}{.5\linewidth}
        \caption{\label{tab:ablation_cot}AUC of \modelnameA{} without and with CoT prompt augmentation.}
        \centering
        \begin{tabular}{p{1.3cm}||c|c}
        \toprule
          & 
        Remove CoT prompting & \modelnameA{}\\
        \midrule
        GSM8K  & 0.837 & 0.951   \\
        CSQA  & 0.665 & 0.769 \\
        SVAMP   & 0.882 & 0.927 \\
        TriviaQA  & 0.792 & 0.817 \\
        \bottomrule
        \end{tabular}
        \end{subtable} 
\end{table}

\subsubsection{Effect of different sentence similarity metrics}
Table \ref{tab:ablation_sim} shows the AUC performance with different similarity metrics. We compare \textbf{Jaccard similarity} calculated by dividing the number of observations in both output strings by the number of observations in either string, \textbf{LLM-embedding} utilizing text-embedding-ada-002\footnote{https://platform.openai.com/docs/api-reference/embeddings} to get embedding for each output answers and calculating the cosine similarities between them, \textbf{NLI} using an off-the-shelf DeBERTa-large model \citep{he2020deberta} for the purpose of categorizing into one of: entailment, contradiction, and neutral, NLI (1-contradiction) using $1-p_{contradiction}$ as the final similarities metrics. Table \ref{tab:ablation_sim} shows that the similarity metric used in \modelnameA{} is essential for discerning the differences among various responses.

\begin{table}[ht]
\caption{Effect of different sentence similarity metrics}
\label{tab:ablation_sim}
\begin{center}
\begin{tabular}{p{1.2cm}||c|c|c|c}
\toprule
Dataset & Jaccard &LLM-embedding  & NLI (1-contradiction) & \modelnameA{}\\
\midrule
GSM8K & 0.896 & 0.866  & 0.892 & 0.951 \\
CSQA  & 0.857 & 0.849  & 0.727 & 0.769 \\
SVAMP & 0.917 & 0.888  & 0.901 & 0.927\\
TriviaQA & 0.650 & 0.642 &  0.794 & 0.817\\
\bottomrule
\end{tabular}
\end{center}
\end{table}

\end{document}